\documentclass[10pt,twocolumn,letterpaper]{article}

\usepackage{cvpr}
\usepackage{times}
\usepackage{epsfig}
\usepackage{graphicx}
\usepackage{amsmath}
\usepackage{amssymb}
\usepackage{subfiles}
\usepackage[numbers,sort,compress]{natbib}
\usepackage{booktabs}
\usepackage{multirow}
\usepackage[toc,page,titletoc]{appendix}
\usepackage{authblk}

\usepackage[pagebackref=true,breaklinks=true,letterpaper=true,colorlinks,bookmarks=false]{hyperref}

\cvprfinalcopy

\ifcvprfinal\fi
\begin{document}

\title{Towards Real-Time Automatic Portrait Matting on Mobile Devices}

\newcommand*\samethanks[1][\value{footnote}]{\footnotemark[#1]}

\author[1]{
Seokjun Seo\thanks{Equal contributions.}
}
\author[1]{
Seungwoo Choi\samethanks[1]
}
\author[1]{
Martin Kersner\samethanks[1]
}
\author[1]{
Beomjun Shin\samethanks[1]
}
\author[2]{
Hyungsuk Yoon\thanks{Work done while at Hyperconnect.}
}
\author[1]{
Hyeongmin Byun
}
\author[1]{
Sungjoo Ha
}
\affil[1]{Hyperconnect\\
Seoul, South Korea}
\affil[2]{Unaffiliated\\
Seoul, South Korea}
\affil[ ]{\tt \{seokjun.seo, seungwoo.choi, martin.kersner\}@hpcnt.com}
\affil[ ]{\tt youhanmir@gmail.com}
\affil[ ]{\tt \{hyeongmin.byun, shurain\}hpcnt.com}

\maketitle

\begin{abstract}
We tackle the problem of automatic portrait matting on mobile devices.
The proposed model is aimed at attaining real-time inference on mobile devices with minimal degradation of model performance.
Our model MMNet, based on multi-branch dilated convolution with linear bottleneck blocks, outperforms the state-of-the-art model and is orders of magnitude faster.
The model can be accelerated four times to attain 30 FPS on Xiaomi Mi 5 device with moderate increase in the gradient error.
Under the same conditions, our model has an order of magnitude less number of parameters and is faster than Mobile DeepLabv3 while maintaining comparable performance.
The accompanied implementation can be found at \url{https://github.com/hyperconnect/MMNet}.
\end{abstract}

\section{Introduction}
\begin{figure}[t]
\begin{center}
    \includegraphics[width=0.95\linewidth]{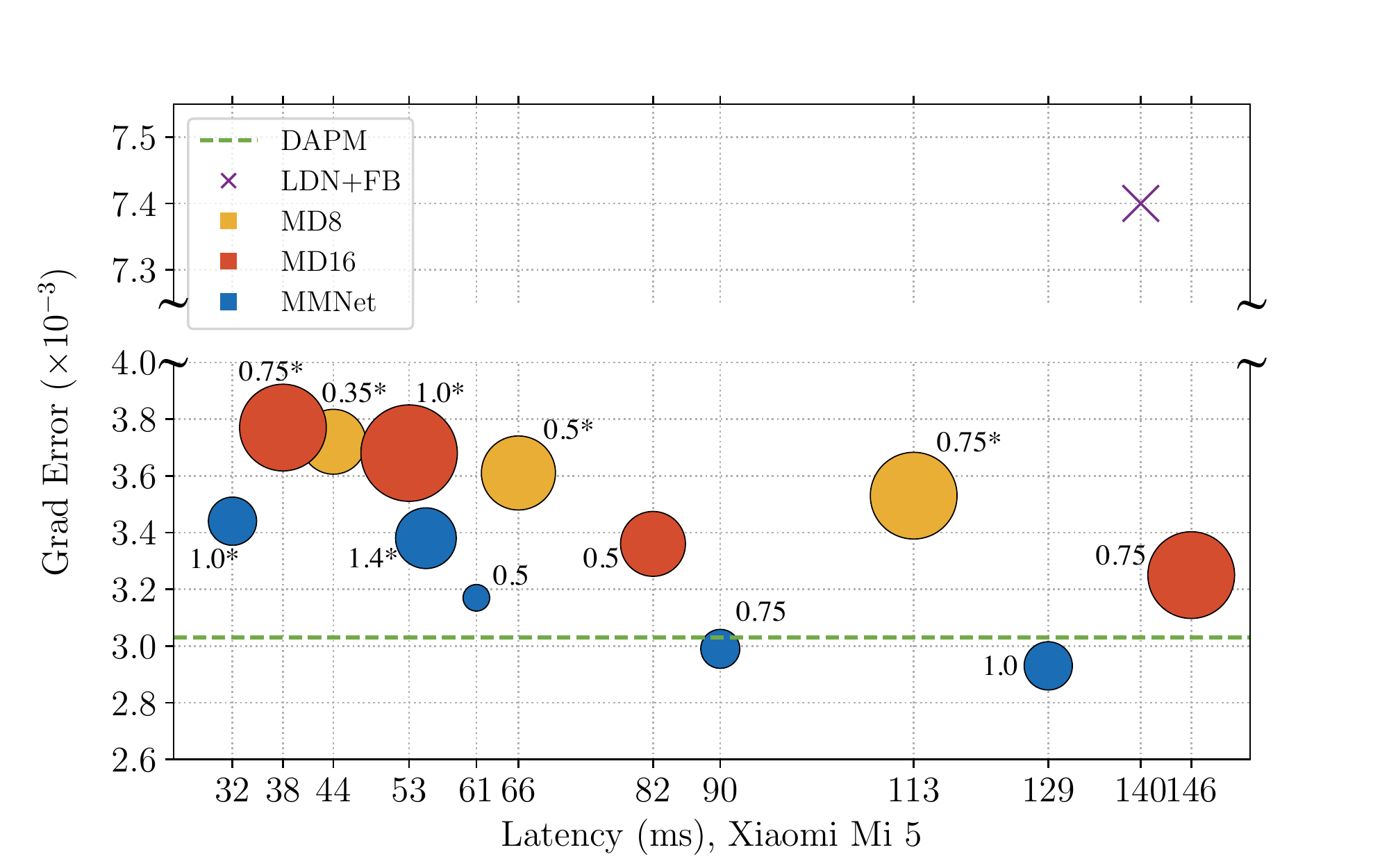}
\end{center}
\caption{
    The trade-off between gradient error and latency on a mobile device.
    Latency is measured using a Qualcomm Snapdragon 820 MSM8996 CPU.
    Size of each circle is proportional to the logarithm of the number of parameters used by the model.
    Different circles of Mobile DeepLabv3 are created by varying the output stride and width multiplier.
    The circles are marked with their width multiplier.
    Results using $128 \times 128$ inputs are marked with $*$, otherwise, inputs are in $256 \times 256$.
    Notice that MMNet outperforms all other models forming a Pareto front.
    The number of parameters for LDN+FB is not reported in their paper.
    Best viewed in color.
}
\label{fig:grad-pareto}
\end{figure}

\begin{figure*}[ht!]
\begin{center}
    \includegraphics[width=0.95\linewidth]{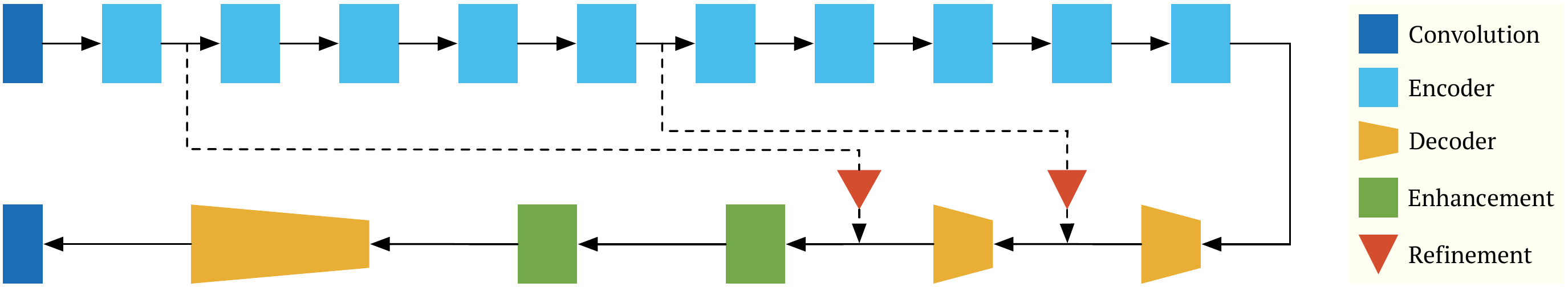}
\end{center}
\caption{The overall structure of the proposed model.
    A standard encoder-decoder architecture is adopted.
    Successively applying encoder blocks summarize spatial information and capture higher semantic information.
    Decoding phase upsamples the image with decoder blocks and improves the result with enhancement blocks.
    Information from skip connections is concatenated with the upsampled information.
    Images are resized to target size before going through the network.
    The resulting alpha matte is converted back to its original resolution.
}
\label{fig:network-structure}
\end{figure*}

Image matting, the task which predicts alpha values of foreground on every pixel, has been studied~\cite{chuang-cvpr-2001, gastal-cgf-2010, he-cvpr-2011, chen-pami-2012-knn-matting, grady-viip-2005}.
Image matting system offers an opportunity for wide applications in computer vision such as color transformation, stylization, and background edits.
It is well-known, however, that image matting is an ill-posed problem~\cite{levin-pami-2008} since seven unknown values (three for foreground RGB, three for background RGB and one for alpha) should be inferred from three known RGB values.
The most widely used method to alleviate the difficulties of the matting problem is to utilize an additional input which roughly separates an image such as trimap~\cite{chuang-cvpr-2001, sun-tog-2004} and scribbles~\cite{levin-pami-2008}.
A trimap splits an image into three parts: definite foreground, definite background, and ambiguous blended regions.
Scribbles, on the other hand, indicate foreground and background with a few strokes.
Even though some of the traditional methods~\cite{shahrian-cvpr-2013, wang-cvpr-2007, levin-pami-2008, sun-tog-2004} work well if additional inputs are provided, it is hard to extend these methods to various image and video matting applications which require real-time performance due to their high computational complexity as well as the dependency on user-interactive inputs.

Other approaches have been studied to automate matting by specifying the object which has to be selected as a foreground~\cite{shen-eccv-2016-dapm, xu-cvpr-2017-dim, chen-acmmm-2018-shm}, for example, portrait matting.
Automatic portrait matting showed even better result than the other methods using trimap~\cite{shen-eccv-2016-dapm}, but the latency is far too high to be used in a real-time application.
\citet{zhu-acmmm-2017} released a lightweight model which can perform automatic matting relatively fast on mobile devices, attaining the latency of 62 ms per image on Xiaomi Mi 5.
However, the gradient error of lightweight model was more than two times worse than that of the state-of-the-art, which made it less attractive in real-world applications.

In this paper, we propose a compact neural network model for automatic portrait matting which is fast enough to run on mobile devices.
The proposed model adopts an encoder-decoder network structure~\cite{badrinarayanan-pami-2015-segnet} and focuses on devising efficient components of the network.
We apply depthwise convolution~\cite{szegedy-cvpr-2015-inception} as the basic convolution operation to extract and downsample features.
The depthwise convolution is considerably cheaper than other convolutions even if we take efficient convolutions such as $1 \times 1$ convolution~\cite{howard-arxiv-2017-mobilenet} into account as well.
The linear bottleneck structure~\cite{sandler-cvpr-2018-mobilenetv2} benefits from the efficiency of depthwise convolutions, boosting the performance while maintaining the latency.
Building upon these observations, the encoder block of the proposed model, consists of multi-branch dilated convolution with linear bottleneck blocks which can reduce the model size with the linear bottleneck structure while aggregating multi-scale information with multi-branch dilated convolutions.
We introduce the width multiplier, a global variable which enlarges or shrinks the number of channels of a convolution, to control the trade-off between the size and the latency of the model.
We incorporate multiple losses into our loss function, including a gradient loss which we propose.

The proposed model shows better performance than the state-of-the-art method while achieving 30 FPS on iPhone 8 without GPU acceleration.
We also evaluate the trade-offs between performance, the latency on mobile devices, and the size of the model.
Our model can achieve 30 FPS on Google Pixel 1 and Xiaomi Mi 5 using a single core while suffering roughly 10\% degradation of gradient error compared to the state-of-the-art.

Our contributions are as follows:
\begin{itemize}
    \item We propose a compact network architecture for automatic portrait matting task which achieves a real-time latency on mobile devices.
    \item We explore multiple combinations of input resolution and width multiplier, which can beat strong baselines for automatic portrait matting on mobile devices.
    \item We demonstrate the capability of each component of the model, including the multi-branch dilated convolution with linear bottleneck blocks, the skip connection refinement block and the enhancement block, through ablation studies.
\end{itemize}

\section{Methods}
Image matting problems take input image $I$, which is a mixture of the foreground image $F$ and the background image $B$.
Each pixel at the $i$-th position is computed as follows:
\begin{equation}
    \label{eq:matting}
    I_i = \alpha_i F_i + (1-\alpha_{i}) B_i,
\end{equation}
where the foreground opacity determines $\alpha_i$.
Since all the quantities on the right-hand side of the Equation~\ref{eq:matting} are unknown, the problem is ill-posed.

%
However, we add an assumption that $F_i$ and $B_i$ are identical to $I_i$ in order to reduce the complexity of the problem.
Even though the assumption may decrease the performance substantially, the empirical result of our experiments show this assumption is reasonable considering the latency gain we get.

\subsection{Model Architecture}

Our model follows a standard encoder-decoder architecture that is widely used in semantic segmentation tasks~\cite{long-cvpr-2015-fcn, ronneberger-miccai-2015-unet, badrinarayanan-pami-2015-segnet}.
Encoder successively reduces the size of the input by downsampling and summarizes the spatial information while capturing higher semantic information.
Decoder, in turn, upsamples the image to recover the detailed spatial information and restores the original input resolution.
The whole network structure of our model, mobile matting network (MMNet), is depicted in Figure~\ref{fig:network-structure}.

Many modern neural network architectures replace a regular convolution with a combination of several cheaper convolutions~\cite{chollet2017xception, wang2017factorized, szegedy2016rethinking}.
Depthwise separable convolution~\cite{howard-arxiv-2017-mobilenet, chollet2017xception} is one of the examples which consists of a depthwise convolution, applying a single convolutional filter per input channel, and a pointwise convolution ($1 \times 1$ convolution) that accumulates the results.
We not only use depthwise separable convolution for some blocks but also adopt the concept of depthwise separable convolution when designing our encoder block.
Depthwise convolution is one such example which we use extensively.
All convolution operations are followed by a batch normalization and a ReLU6 non-linearity except the linear projection operation that is placed at the end of the encoder block~\cite{sandler-cvpr-2018-mobilenetv2}.
Due to the linear bottleneck structure, the information flow from an encoder block to another is projected to a low-dimensional representation.

In the encoder block, the information flowing from the lower layers is expanded by the first $1 \times 1$ multi-branch convolutions.
The linear bottleneck compresses the processed image.
The data upsampled by the decoder block is concatenated with the refined knowledge through a skip connection.
The number of channels for each path are maintained to have the same value.
Table~\ref{tab:model-architecture} details how much each component expands and compresses the information flow.

To control the trade-off between model size and model performance, we adopt width multiplier~\cite{howard-arxiv-2017-mobilenet}.
The width multiplier $\alpha$, is a global hyperparameter that is multiplied to the number of input and output channels to make the layers thinner or thicker depending on the computational budget.

\begin{table}[t]
    \begin{center}
        \begin{tabular}{lcc}
            \toprule



            Name          & Component Details            & Output Size \\
            \midrule
            Initial Block & Conv $3\times 3$, S$2$       & $128 \times 128$, 32 \\
            Encoder 1     & DR $[1, 2, 4, 8]$, S$2$      & $64 \times 64$, 16   \\
            Encoder 2     & DR $[1, 2, 4, 8]$, S$1$      & $64 \times 64$, 24   \\
            Encoder 3     & DR $[1, 2, 4, 8]$, S$1$      & $64 \times 64$, 24   \\
            Encoder 4     & DR $[1, 2, 4, 8]$, S$1$      & $64 \times 64$, 24   \\
            Encoder 5     & DR $[1, 2, 4]$, S$2$         & $32 \times 32$, 40   \\
            Encoder 6     & DR $[1, 2, 4]$, S$1$         & $32 \times 32$, 40   \\
            Encoder 7     & DR $[1, 2, 4]$, S$1$         & $32 \times 32$, 40   \\
            Encoder 8     & DR $[1, 2, 4]$, S$1$         & $32 \times 32$, 40   \\
            Encoder 9     & DR $[1, 2]$, S$2$            & $16 \times 16$, 80   \\
            Encoder 10    & DR $[1, 2]$, S$1$            & $16 \times 16$, 80   \\
            Decoder 1     & Upsample $\times 2$ (Skip 5) & $32 \times 32$, 128  \\
            Decoder 2     & Upsample $\times 2$ (Skip 1) & $64 \times 64$, 80   \\
            Enhancement 1 & DR $[1, 2, 4]$, S$1$         & $64 \times 64$, 40   \\
            Enhancement 2 & DR $[1, 2, 4]$, S$1$         & $64 \times 64$, 40   \\
            Decoder 3     & Upsample $\times 4$          & $256 \times 256$, 16 \\
            Final Block   & Conv $1\times 1$, Softmax    & $256 \times 256$, 2  \\
            \bottomrule
        \end{tabular}
    \end{center}
    \caption{
        The model architecture of MMNet.
        We assume that width multiplier are set to $1.0$.
        Decoder \#1 and \#2 are connected to encoder \#5 and \#1 with a skip connection and a refinement block, respectively.
        DR denotes the dilation rates in the multi-branch dilated convolutions.
        S represents the stride value in the strided convolution.
    }
    \label{tab:model-architecture}
\end{table}

\subsubsection{Encoder Block}

\begin{figure}[b]
\begin{center}
    \includegraphics[width=0.95\linewidth]{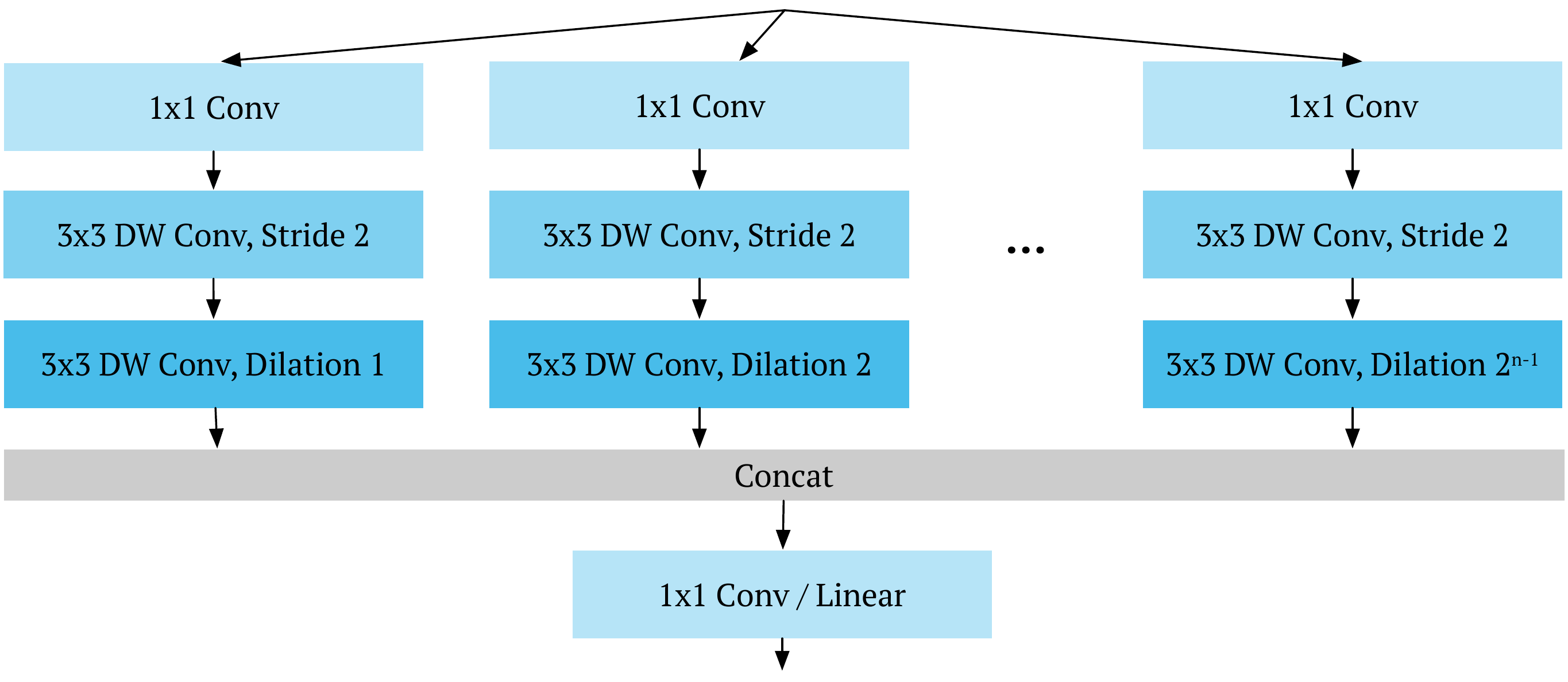}
\end{center}
\caption{The encoder block.
    It employs a multi-branched dilated convolution with a linear bottleneck.
    The linear bottleneck compresses the information to a low-dimensional representation before handing it over to the next encoder block.
}
\label{fig:encoder-block}
\end{figure}

MMNet encoder block has a multi-branched dilated convolution structure with a linear bottleneck.
Input flows to multiple branches which undergo channel expansion followed by a strided convolution and a dilated convolution.
The dilation rates are different for all branches following $2^{n-1}$ rates.
Multi-branch dilated convolution amounts to sampling spatial information at different scales.
The outputs of different branches are concatenated to form a tensor containing multi-scale information.
Applying encoder blocks in succession allows the network to capture multi-level information increasingly.
As the encoder blocks are consecutively applied, we decrease the number of branches in an encoder block, slowly changing the dilation rates from $[1, 2, 4, 8]$ to $[1, 2]$.

A linear bottleneck structure is imposed on the encoder block where the output of the encoder block is thinner than the intermediate representations.
The final convolution after combining the multi-branch information projects the input to a low-dimensional compressed representation.
The linear bottleneck is a decomposition of a regular convolution that connects two encoder blocks into two cheaper convolutions with reduced channels.
The encoder block is illustrated in Figure~\ref{fig:encoder-block}.

\subsubsection{Decoder Block}

\begin{figure}[t]
\begin{center}
    \includegraphics[width=0.975\linewidth]{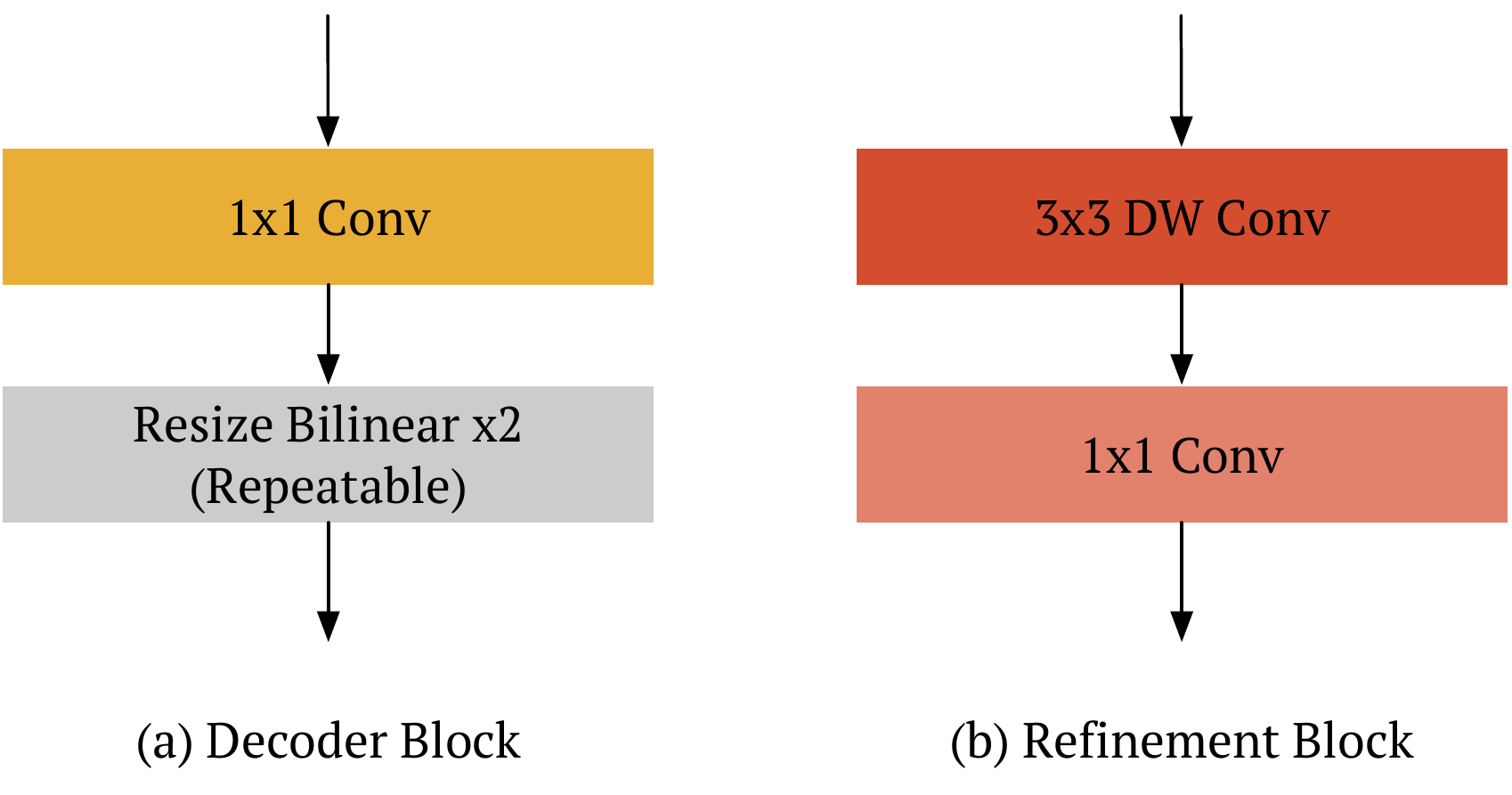}
\end{center}
\caption{
The decoder block (a) upsamples bilinearly which could be repeated multiple times to upsample by a larger factor.
The refinement block (b) is added to each skip connection where the direct information from a lower level is refined before merging with the higher level information from a decoder block.
}
\label{fig:decoder-refinement-block}
\end{figure}

The decoder performs multiple upsampling to restore the initial resolution of the input image.
To help decoder with the restoration of low-level features from compressed spatial information, skip connections are employed to directly connect the output of the lower-layer encoder to its corresponding decoder~\cite{ronneberger-miccai-2015-unet}.
Instead of using the information provided by the corresponding encoder blocks without any modifications we refine the information by performing a $3 \times 3$ depthwise separable convolution.
The resulting refined information is concatenated with the upsampled information.
This specific refinement technique is reminiscent of the refinement module proposed in SharpMask~\cite{pinheiro-eccv-2016-refinemodule1, treml-nips-2016-refinemodule2}.
A decoder block with a refinement block is illustrated in Figure~\ref{fig:decoder-refinement-block}.
In this work, we connect the feature map of encoder \#1 and encoder \#5 to decoder \#2 and decoder \#1, respectively.
In the final decoder block, we perform $4\times$ upsampling instead of the usual $2\times$ to shorten the decoding pipeline.

\subsubsection{Enhancement Block}

As the decoder block keeps upsampling the feature map, there is no way to enhance the predictions of neighboring values.
To tackle this problem, we insert two enhancement blocks in the middle of the decoding phase.
Rather than designing a new block, we share the same architecture with encoder block.
The only difference between enhancement block and encoder block is that depthwise convolution with stride two is removed because the enhancement block should sustain the resolution of a feature map.
In the ablation study, we show the effectiveness of the enhancement block.

\subsection{Loss Functions}

The alpha loss and the compositional loss are frequently used in matting tasks.
The alpha loss $L_{\alpha}$, measures the mean absolute difference between the ground truth mask and the mask predicted by the model.
The compositional loss $L_{\text{c}}$, measures the mean absolute difference between the values of ground truth RGB foreground pixels and the model predicted RGB foreground pixels.
The compositional loss penalizes the model when the model incorrectly predicts a pixel with high value.
\begin{align}
\label{eq:2}
L_{\alpha} &= \frac{1}{K} \sum_{i=1}^{K} {\left\lvert \alpha_i - \alpha_i^{\text{gt}} \right\rvert} \\
L_{\text{c}} &= \frac{1}{3K} \sum_{i=1}^{K} \sum_{j=1}^{3} {\left\lvert \alpha_{ij} I_{ij} - \alpha_{ij}^{\text{gt}} I_{ij} \right\rvert},
\end{align}
where the $K$ is equal to the width time height, $W \times H$, and $\alpha$ is a vectorized alpha matte where each pixel value is indexed by subscript $i$.
The $\text{gt}$ superscript denotes the alpha matte is from ground truth.

We use the KL divergence between the ground truth $A^{\text{gt}} \in \mathbb{R}^{W \times H}$ and the model predicted $A \in \mathbb{R}^{W \times H}$.
The KL divergence is defined to be:
\begin{align}
\label{eq:kl}
L_{\text{KL}} &= -p(A^{\text{gt}}) \log \frac{p(A)}{p(A^{\text{gt}})} \\
&= -p(A^{\text{gt}}) \log p(A) + p(A^{\text{gt}}) \log p(A^{\text{gt}}).
\end{align}
The second term is the entropy of the ground truth alpha matte, which is constant with respect to model predicted $A$.
Removing the second term leads to optimization of the following loss:
\begin{equation}
    \tilde{L}_{\text{KL}} = \sum_{i=1}^{K} \left( \alpha_i^{\text{gt}} \log \alpha_i + (1 - \alpha_i^{\text{gt}}) \log (1 - \alpha_i) \right).
\end{equation}

Two additional loss terms are included in the loss function.
An auxiliary loss~\cite{szegedy-cvpr-2015-inception} $L_{\text{aux}}$, helps with the gradient flow by including an additional KL divergence loss between the downsampled ground truth mask and the output of the encoder block \#10.
A gradient loss $L_{\text{grad}}$, guides the model to capture fine-grained details in the edges.
We use Sobel-like filter $\mathbf{S}:$
\begin{align}
    \mathbf{S} &= \left[
    \begin{array}{ccc}
        -\frac{1}{8} & 0 & \frac{1}{8} \\
        -\frac{2}{8} & 0 & \frac{2}{8} \\
        -\frac{1}{8} & 0 & \frac{1}{8} \\
    \end{array}
    \right],
\end{align}
to create a concatenation of two image derivatives $\mathbf{G}(A) = [\mathbf{S} * A, \mathbf{S}^T * A]$ where $*$ is a convolution.
The resulting $\mathbf{G}(\cdot)$ yields a two-channel output that contains the gradient information along $x$-axis and $y$-axis.
We apply $\mathbf{G}(\cdot)$ to both the ground truth mask and the model predicted mask to compute the mean absolute differences.
The gradient loss is computed as follows:
\begin{align}
        L_{\text{grad}} &= \frac{1}{K} \sum_{i=1}^K {\left\lvert \mathbf{G}(A)_i - \mathbf{G}(A^{\text{gt}})_i \right\rvert} \\
                        &= \frac{1}{2K} \sum_{i=1}^K \left(\left\lvert (\mathbf{S} * A - \mathbf{S} * A^{\text{gt}} )_i\right\rvert \right. \nonumber\\
                        &\left.\quad\quad\quad\quad\quad\quad + \left\lvert (\mathbf{S}^T * A - \mathbf{S}^T * A^{\text{gt}})_i \right\rvert \right).
\end{align}

The following Equation~\ref{eq:complex-loss} depicts the loss function of our proposed network.
\begin{equation}
    \label{eq:complex-loss}
    L = \beta_1 L_{\alpha} + \beta_2 L_{\text{c}} + \beta_3 \tilde{L}_{\text{KL}} + \beta_4 L_{\text{grad}} + \beta_5 L_{\text{aux}},
\end{equation}
where we set $\beta$ values to control the influence of each loss terms.
We set them to have equal values of one for the following experiments.

\section{Experiments}
Automatic portrait matting takes input image with a portrait and denotes each pixel with a linear mixture of the foreground and the background.

We use data provided by \citet{shen-eccv-2016-dapm} which consists of 2,000 images of $600 \times 800$ resolution where 1,700 and 300 images are split as training and testing set respectively.
To overcome the lack of training data, we augment images by utilizing scaling, rotation and left-right flip.
First, an image is rescaled to the input size of the model and random scaling factor is selected from $1$ to $1.15$.
The image is then scaled with the selected factor.
Rotation by $[-15^{\circ}$, $15^{\circ}]$ is applied with a probability of $0.5$ which means that half of the augmented images are not rotated.
Additional cropping is computed to make the size of the image to match the input size of the model.
Finally, the left-right flip is also applied with a probability of $0.5$.

To train our model, we optimize our proposed model with respect to the loss function in Equation~\ref{eq:complex-loss} using Adam optimizer with a batch size of 32 and a fixed learning rate of $1 \times 10^{-4}$.
Input images were resized to $128 \times 128$ and $256 \times 256$.
The model trained on $128 \times 128$ images are faster but produces worse alpha mattes compared to the model trained on $256 \times 256$ images.
Weight decays were set to $4\times10^{-7}$.
All experiments are conducted using a TensorFlow~\cite{abadi-osdi-2016-tensorflow} trained on a single Titan V GPU.

Following the work of \citet{zhu-acmmm-2017}, we used gradient error to evaluate our model in portrait matting problem.
The gradient error as a metric, which is different from gradient loss, is defined as:
\newcommand{\normone}[1]{\left\lvert#1\right\rvert}
\newcommand{\normtwo}[1]{\left\lVert#1\right\rVert}
\begin{equation}
    \label{eq:grad-metric}
    \frac{1} {K} \sum_{i} \normtwo{\nabla \alpha_i - \nabla \alpha_i^{\text{gt}}},
\end{equation}
where $\alpha$ is the alpha matte predicted by the model, and $\alpha^{\text{gt}}$ is the corresponding ground truth and $K$ is equal to width $\times$ height.
$\nabla$ denotes the differential operator that is computed by convolving the alpha map with first-order Gaussian derivative filters with variance $1.4$~\cite{rhemann-cvpr-2009-grad-loss}.

Another metric we use to evaluate our model is the mean absolute differences (MAD).
The MAD is defined as follows:
\begin{equation}
    \frac{1}{K} \sum_i \normone{\alpha_i - \alpha_i^{\text{gt}}}.
\end{equation}

For a fair comparison with previous methods, we scale the predicted alpha matte to the original size of input images, $600 \times 800$ in this case, and calculate evaluation metrics.

We compare our model to DAPM~\cite{shen-eccv-2016-dapm}, LDN+FB\cite{zhu-acmmm-2017}, and Mobile DeepLabv3~\cite{sandler-cvpr-2018-mobilenetv2}.
Mobile DeepLabv3 exploits MobileNetV2 as its feature extractor and has its atrous spatial pyramid pooling (ASPP) module removed as suggested by \citet{sandler-cvpr-2018-mobilenetv2}.
We use Equation~\ref{eq:complex-loss} to optimize Mobile DeepLabv3 in equal footings as MMNet, but remove the auxiliary loss since it requires a modification to the network architecture.

\section{Results}
\subsection{Matting Performance}

\begin{figure*}[t]
\begin{center}
    \includegraphics[width=0.95\linewidth]{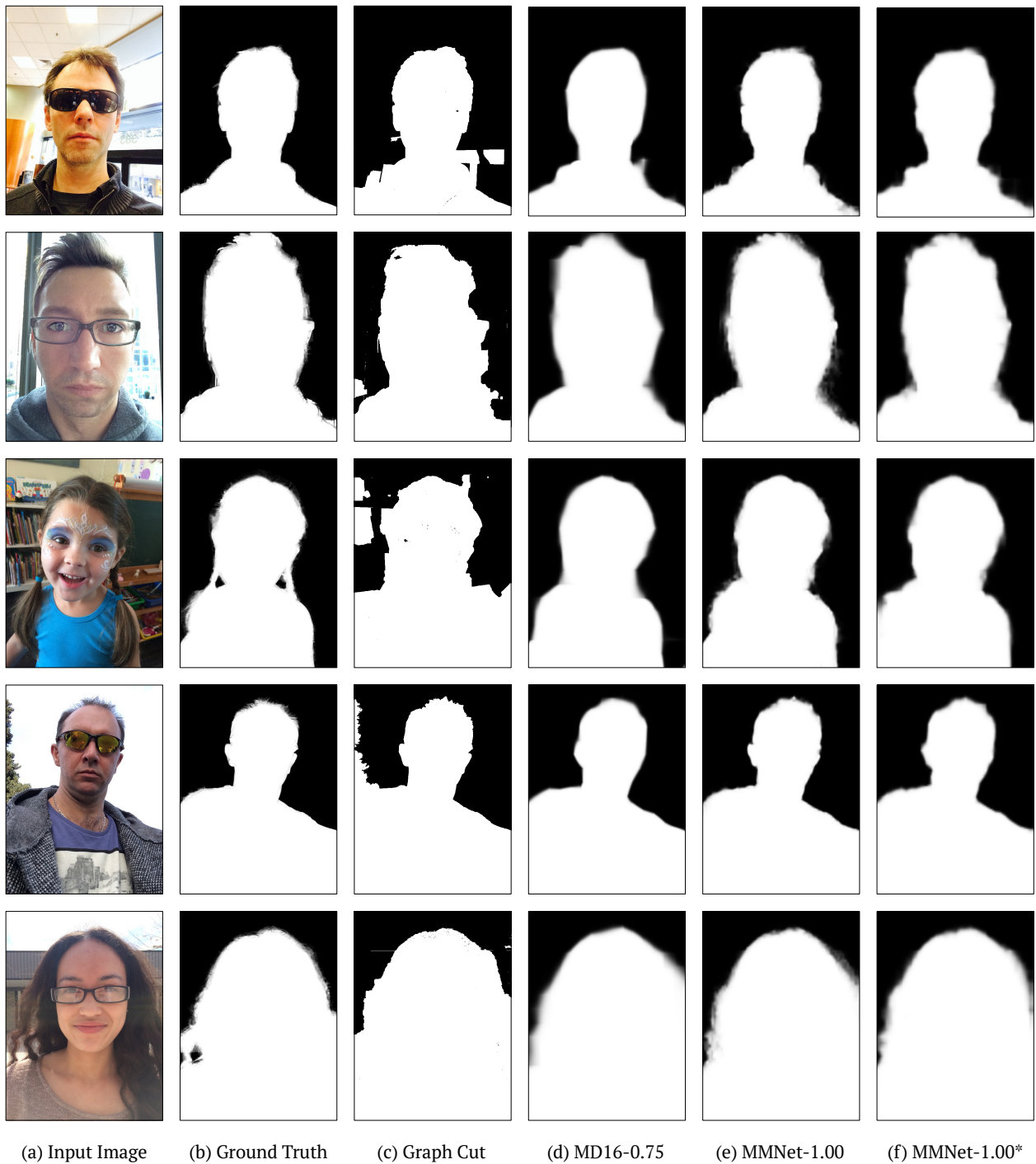}
\end{center}
\caption{
    Visual comparison of different models.
    Graph Cut~\cite{rother-tog-2004-grabcut} results were obtained using OpenCV library~\cite{opencv}.
    The column marked with $*$ displays the result using $128 \time 128$ inputs.
    MMNet is better able to construct delicate details compared to other models.
    Note that MMNet with $128 \times 128$ input still outputs a reasonable alpha matte despite its reduced capacity.
}
\label{fig:visual-comparison}
\end{figure*}

Table~\ref{tab:ldn-fb-compare} compares the result of DAPM~\cite{shen-eccv-2016-dapm}, LDN+FB~\cite{zhu-acmmm-2017}, Mobile DeepLabv3~\cite{sandler-cvpr-2018-mobilenetv2}, and the proposed method.
Input images were scaled to $128 \times 128$ or $256 \times 256$, depending on the hyper-parameter.
When smaller images are fed into the network, the latency drops considerably at the expense of the quality of the alpha matte.
Input images were rescaled back to their original resolutions before evaluation.
The gradient error and the latency for DAPM and LDN+FB were reported by \citet{zhu-acmmm-2017}.
For a fair comparison, we compute the latency of the models on a Xiaomi Mi 5 device (Qualcomm Snapdragon 820 MSM8996 CPU), as suggested by \citet{zhu-acmmm-2017}.
Since \citet{zhu-acmmm-2017} did not report how much CPU resources they used, we measure the latency by restricting the use to a single core.
Specifically, we use TensorFlow Lite~\cite{tflite} benchmark tool to compute the latency of Mobile DeepLabv3 and MMNet by averaging 100 runs of the model inference on a Xiaomi Mi 5 device while restricting the models to use a single thread.

\citet{zhu-acmmm-2017} reports that DAPM takes 6 seconds on a computer with Core E5-2600 @2.60Ghz CPU.
MMNet-1.0 outperforms DAPM while running orders of magnitude faster on a mobile CPU.
When the input image is resized to $128 \times 128$ for faster inference, our model attains real-time inference, surpassing the rate of 30 frames per second.
The real-time version of MMNet is still competitive against DAPM with a moderate increase in its gradient error.

The visual comparison of alpha matte in Figure~\ref{fig:visual-comparison} illustrates the qualitative differences of different models.
MMNet is better able to construct the finer details compared to other models.
Even the real-time version of MMNet produces a reasonable alpha matte regardless of its reduced capacity.

\begin{table}[t]
    \begin{center}
        \begin{tabular}{lrr}
            \toprule
            \multirow{2}{*}{Method}                                                & Time        & Gradient Error     \\
                                                                                   & (ms)        & ($\times 10^{-3}$) \\
            \midrule
            Graph-cut trimap${}^{\dagger}$                                         & -           & 4.93               \\
            Trimap by~\cite{shen-computergraphicsforum-2016-pfcn+}${}^{\dagger}$   & -           & 4.61               \\
            Trimap by FCN~\cite{long-cvpr-2015-fcn}${}^{\dagger}$                  & -           & 4.14               \\
            Trimap by DeepLab~\cite{chen-pami-2018-deeplab}${}^{\dagger}$          & -           & 3.91               \\
            Trimap by CRFasRNN~\cite{zheng-iccv-2015-crfasrnn}${}^{\dagger}$       & -           & 3.56               \\
            DAPM~\cite{shen-eccv-2016-dapm}                                        & -           & 3.03               \\
            LDN+FB~\cite{zhu-acmmm-2017}                                           & 140         & 7.40               \\
            MD16-0.75                                                              & 146         & 3.23               \\
            MD16-1.0                                                               & 203         & 3.22               \\
            MD16-0.75$*$                                                           & 38          & 3.71               \\
            MMNet-1.0                                                              & 129         & 2.93               \\
            MMNet-1.4                                                              & 213         & \textbf{2.86}               \\
            MMNet-1.0$*$                                                           & \textbf{32} & 3.38               \\
            \bottomrule
        \end{tabular}
    \end{center}
    \caption{Model comparisons on the test split.
        Time is computed on Xiaomi Mi 5 phone.
        Mobile DeepLabv3 used output stride of 16.
        Floating point numbers in the method name indicate the width multiplier.
        The row marked with $*$ displays the result using $128 \time 128$ inputs.
        Our model outperforms other models while processing images at a faster rate.
        The experiments marked with $\dagger$ are copied from \citet{shen-eccv-2016-dapm}.
    }
    \label{tab:ldn-fb-compare}
\end{table}

\subsection{Real-Time Inference on Mobile Devices}

To examine the trade-off between execution time and model performance, we explore the model space by varying the width multiplier values and the input resolution.
We compare our model with Mobile DeepLabv3 suggested by \citet{sandler-cvpr-2018-mobilenetv2}.
Table~\ref{tab:mobile-deeplab} details the result of the experiment.
The results are sorted by the latency and models with comparable execution time are clustered using horizontal dividers.
We see that our proposed model dominates Mobile DeepLabv3 in all clusters in terms of gradient error.
Also, note that the number of parameters differs by an order of magnitude.
Requiring a small number of parameters is especially appealing if we target a mobile device since end-users do not have to download a bulky model whenever there is an update of the model.

\begin{table}[t]
    \begin{center}
        \begin{tabular}{lcccr}
            \toprule
            \multirow{2}{*}{Method}        & Time         & Gradient            & MAD           & Params \\
                                           & (ms)         & $(10^{-3})$          & $(10^{-2})$          & (M)    \\
            \midrule
            MD16-0.75                      & 146          & 3.25          & \textbf{2.31} & 1.327  \\
            MMNet-1.00                     & \textbf{129} & \textbf{2.93} & 2.48          & 0.199  \\
            \midrule
            MD8-0.75$*$                    & 113          & 3.53          & 2.61          & 1.327  \\
            MMNet-0.75                     & 90           & \textbf{2.99} & 2.65          & 0.127  \\
            MD16-0.50                      & 82           & 3.36          & \textbf{2.53} & 0.454  \\
            MD8-0.50$*$                    & 66           & 3.61          & 2.85          & 0.713  \\
            MMNet-0.50                     & \textbf{61}  & 3.17          & 2.83          & 0.069  \\
            \midrule
            MMNet-1.40$*$                  & 55           & \textbf{3.38} & \textbf{2.72} & 0.369  \\
            MD16-1.00$*$                   & 53           & 3.68          & 2.88          & 2.142  \\
            MD8-0.35$*$                    & 44           & 3.72          & 3.07          & 0.454  \\
            MD16-0.75$*$                   & 38           & 3.77          & 2.96          & 1.327  \\
            MMNet-1.00$*$                  & \textbf{32}  & 3.44          & 2.97          & 0.199  \\
            \midrule
            \midrule
            MMNet-1.00Q                    & 98           & 2.88          & 2.47          & 0.199  \\
            \bottomrule
        \end{tabular}
    \end{center}
    \caption{
        Comparison of MMNet against Mobile DeepLabv3.
        Floating point numbers in the method name indicate the width multiplier.
        The row marked with $*$ displays the result using $128 \time 128$ inputs.
        Output strides of 8 and 16 were tested for Mobile DeepLabv3.
        Note that the proposed model dominates Mobile DeepLabv3 when the latency is less than 60.
        In slower regime, MMNet still outperforms Mobile DeepLabv3 in gradient error but are sometimes worse in MAD.
        Quantized model is included in the last row.
    }
    \label{tab:mobile-deeplab}
\end{table}

Figure~\ref{fig:grad-pareto} plots trade-off between gradient error and latency on a mobile device.
Note that MMNet develops a Pareto-front in this space and outperforms other models.
Latency comparison of Pixel 1 and iPhone 8 are included in the supplementary material.

\subsection{Ablation Studies}

Our proposed network owes its performance to several building blocks utilized in its model architecture.
We analyze the impact of each design choices by performing ablation experiments.

\subsubsection{Network Component}

\paragraph{Dilation Rates in Encoder Block}
We study the effect of different dilation rates in the encoder block.
The proposed model contains a multi-branch dilated convolutions in the encoder block.
We analyze the impact of this decision by fixing the dilation rates to one.

\paragraph{Refinement Block}
Whenever there is a skip connection, we have included a refinement block to improve the decoding quality.
The refinement block enhances the result of the encoder block by performing $3 \times 3$ depthwise separable convolution followed by batch normalization and a ReLU6 non-linearity.
We remove the refinement block and study its impact on the final result.

\paragraph{Enhancement Block}
The enhancement blocks are intended to give the network a layer to improve the final result before its resolutions are fully recovered.
We study the effect of the enhancement block by removing it entirely from the network.

\begin{table}[t]
    \begin{center}
        \begin{tabular}{lc}
            \toprule
            \multirow{2}{*}{Method}           & Gradient Error     \\
                                              & ($\times 10^{-3}$) \\
            \midrule
            No dilation                       & 3.25               \\
            No enhancement in decoding        & 3.04               \\
            No refinement in skip connection  & 3.07               \\
            Proposed model                    & 2.93               \\
            \bottomrule
        \end{tabular}
    \end{center}
    \caption{Ablation study on the test split of matting dataset.
        All experiments are performed using MMNet with width multiplier of 1.0.
    }
    \label{tab:ablation-architecture}
\end{table}

Table~\ref{tab:ablation-architecture} illustrates the results when different components of the model architecture are modified.
We see that all the components contribute to the final performance of the proposed model.
When the dilation rate is fixed to one, the network has a hard time generalizing due to its limited effective receptive field.
Enhancement and refinement in the decoding phase also boost the network performance.

\subsection{Quantization}
We demonstrate the full pipeline for training a real-time portrait matting model targeting a mobile platform by incorporating quantization of our model.
Quantization of model parameters and its activation reduces the bit-width required by the model.
The reduction of bit-width allows one to exploit integer arithmetics in boosting the network inference speed.

The target model undergoes a quantization-aware training phase via fake quantization~\cite{jacob-cvpr-2018-quantization}.
While maintaining full precision weights, tensors are downcasted to fewer bits during the forward pass.
On a backward pass, the full precision weights are updated instead of downcasted tensors from which the gradients are computed.
Once the training is complete, quantized models are executed using the TensorFlow Lite framework~\cite{tflite}.

Table~\ref{tab:mobile-deeplab} contains the result of 8-bit quantized model.
The model enjoys 25\% decrease in latency and better gradient error.
The details for quantization are included in the supplementary material.

\section{Related Work}
Image matting task has been mostly approached using sampling~\cite{chuang-cvpr-2001, gastal-cgf-2010, he-cvpr-2011, shahrian-cvpr-2013, wang-cvpr-2007} or propagation-based~\cite{chen-pami-2012-knn-matting, grady-viip-2005, levin-pami-2008, sun-tog-2004} ideas.
Recently, with the success of convolutional neural networks (CNN) in computer vision tasks, there has been a growing number of works utilizing CNNs.
\citet{cho-eccv-2016} proposed end-to-end network which relies on other matting algorithms' outputs, such as the closed form matting~\cite{levin-pami-2008} and the KNN matting~\cite{chen-pami-2012-knn-matting}, to produce the final alpha matte.
\citet{shen-eccv-2016-dapm} proposed an automatic image matting method leveraging CNN to create a trimap which is fed to closed form matting~\cite{levin-pami-2008} by backpropagating the matting error back to the trimap network.
\citet{xu-cvpr-2017-dim} take the approach further by directly learning the alpha matte.
\citet{chen-acmmm-2018-shm} combine trimap generation and alpha matte generation using a fusion module.

Many works on image matting are mainly focused on achieving higher accuracy rather than the real-time inference of models.
But recently, researchers are shifting the focus to networks that accommodate real-time inference~\cite{zhu-acmmm-2017}.
\citet{zhu-acmmm-2017} studied real-time portrait matting on mobile devices which is directly comparable to our result.

Since the work of \citet{long-cvpr-2015-fcn}, fully convolutional networks (FCN) have been widely used in various segmentation tasks~\cite{zheng-iccv-2015-crf-rnn, jegou-cvprw-2017-tiramisu}.
Many of the semantic segmentation networks adopt encoder-decoder structure~\cite{badrinarayanan-pami-2015-segnet}.
The proposed model uses skip connections to concatenate the output of an encoder block to a decoder block which has been known to improve the result of semantic pixel-wise segmentation tasks~\cite{ronneberger-miccai-2015-unet}.

\citet{chen-pami-2018-deeplab} proposed DeepLab~\cite{chen-arxiv-2017-deeplabv3, chen-eccv-2018-deeplabv3+} architecture which extensively uses the ASPP module.
ASPP module aims to solve the problem of efficient upsampling and handling objects at multiple scales.
Our model adopts a multi-branch structure from Inception network~\cite{szegedy-cvpr-2015-inception}, together with the dilated convolution of different dilation rates, which resembles the ASPP module.

One of the most prominent light-weight neural networks is MobileNet and its variants~\cite{howard-arxiv-2017-mobilenet, sandler-cvpr-2018-mobilenetv2}.
Depthwise separable convolution was shown to be extremely effective in creating a light-weight network while keeping the accuracy drop to a tolerable level.

ENet, an efficient neural network architecture designed with the intention of tackling a semantic segmentation task, was proposed by \citet{paszke-arxiv-2016-enet}.
Our work is inspired by the design choices detailed in their work for creating an efficient neural network.

\section{Conclusions}
In this work, we have proposed an efficient model for performing automatic portrait matting task on mobile devices.
We were able to accelerate the model four times to achieve 30 FPS on Xiaomi Mi 5 device with only 15\% increase in the gradient error.
Comparison against Mobile DeepLabv3 showed that our model is not only faster when the performance is comparable, but also requires an order of magnitude less number of parameters.
Through ablation studies, we have shown that our choice of the multi-branch dilated convolution with a linear bottleneck is essential in maintaining high performance.
We also make our implementation available at \url{https://github.com/hyperconnect/MMNet}.

A general extension of our work is to handle general image matting problem, such as automatic saliency matting.
Since we can already achieve real-time, it is natural to extend the work further by tackling the video matting problem as well.
Pushing for real-time inference on mobile devices requires a carefully prepared pipeline for it to work in a real-world setting.
Distillation to guide the mobile-friendly model in training and even lower-bit quantization for added speedup is highly desired.

{\small
\setlength{\bibsep}{0pt}
\bibliographystyle{abbrvnat}
\bibliography{longstrings,segmentation}
}

\appendixpageoff
\appendixtitleoff
\renewcommand{\appendixtocname}{Supplementary Material}
\begin{appendices}
    \begin{center}
    \textbf{\large Supplemental Materials}
    \end{center}

    \section{Quantization}
We used \verb|tensorflow.contrib.quantize| to quantize our model.
Custom implementation of \verb|resize_bilinear| operation, optimized using SIMD instructions was deployed.
Since we are using fake quantization~\cite{jacob-cvpr-2018-quantization} for quantization-aware training, additional fake quantization node was inserted after a \verb|resize_bilinear| operation.

The quantized version of softmax provided by TensorFlow Lite is slow for our use case since it is optimized for a classification task.
Our formulation allows us the make an assumption that the output has only two channels.
Quantizing the values to 8-bits means that there are only 65,536 valid logit pairs.
Instead of explicit computation of softmax, we precompute the values and substitute the calculation with a table lookup.

    \section{Latency}
\begin{table}[h!]
    \begin{center}
        \begin{tabular}{lccccc}
            \toprule
            Method              & Pixel 1     & Mi 5        & iPhone 8   \\
            \midrule
            MD16-0.75           & $142 \pm 2$ & $146 \pm 1$ & $32 \pm 0$ \\
            MMNet-1.00          & $127 \pm 1$ & $129 \pm 1$ & $30 \pm 0$ \\
            MD8-0.75$*$         & $111 \pm 1$ & $113 \pm 1$ & $22 \pm 0$ \\
            MMNet-0.75          & $87 \pm 1$  & $90 \pm 1$  & $22 \pm 0$ \\
            MD16-0.50           & $80 \pm 1$  & $82 \pm 1$  & $19 \pm 0$ \\
            MD8-0.50$*$         & $66 \pm 1$  & $66 \pm 1$  & $13 \pm 0$ \\
            MMNet-0.50          & $60 \pm 0$  & $61 \pm 1$  & $15 \pm 0$ \\
            MMNet-1.40$*$       & $53 \pm 1$  & $55 \pm 1$  & $12 \pm 0$ \\
            MD16-1.00$*$        & $53 \pm 1$  & $53 \pm 1$  & $12 \pm 0$ \\
            MD8-0.35$*$         & $45 \pm 1$  & $44 \pm 0$  & $9 \pm 0$  \\
            MD16-0.75$*$        & $38 \pm 1$  & $38 \pm 1$  & $8 \pm 0$  \\
            MMNet-1.00$*$       & $33 \pm 1$  & $32 \pm 1$  & $8 \pm 0$  \\
            \midrule
            \midrule
            MD16-0.75           & $104 \pm 1$ & -           & -          \\
            MMNet-1.00Q         & $98 \pm 2$  & $98 \pm 1$  & -          \\
            \bottomrule
        \end{tabular}
    \end{center}
    \caption{
        Latency of models on different mobile devices.
        All numbers are in milliseconds.
        The row marked with $*$ displays the result using $128 \time 128$ inputs.
        Quantized model is included in the last row.
    }
    \label{tab:latency}
\end{table}

Table~\ref{tab:latency} depicts the latency of different models measured on Pixel 1, Xiaomi Mi 5, and iPhone 8.
All measurements are performed with the TensorFlow Lite~\cite{tflite} benchmark tool on a mobile device while restricting the models to use a single thread.
The mean and the standard deviation obtained from 100 runs are included in the table.
The measurements were separated apart in time to give the device enough time to cool down.
Demo video is available at \url{https://github.com/hyperconnect/MMNet}.

    \section{Detailed Architectures}
\begin{table*}[b]
    \begin{center}
        \begin{tabular}{lcccccc}
            \toprule
            Name                 & \multicolumn{4}{c}{Output channels of 1x1 convolution}    \\
                                 & First & Encoder/Enhancement & Decoder & Refinement & Final \\
            \midrule
            Initial Block        & $32$ & $-, -$              & $-$     & $-$        & $-$   \\
            Encoder 1            & $-$  & $16, 16$            & $-$     & $-$        & $-$   \\
            Encoder 2            & $-$  & $16, 24$            & $-$     & $-$        & $-$   \\
            Encoder 3            & $-$  & $24, 24$            & $-$     & $-$        & $-$   \\
            Encoder 4            & $-$  & $24, 24$            & $-$     & $-$        & $-$   \\
            Encoder 5            & $-$  & $32, 40$            & $-$     & $-$        & $-$   \\
            Encoder 6            & $-$  & $64, 40$            & $-$     & $-$        & $-$   \\
            Encoder 7            & $-$  & $64, 40$            & $-$     & $-$        & $-$   \\
            Encoder 8            & $-$  & $64, 40$            & $-$     & $-$        & $-$   \\
            Encoder 9            & $-$  & $80, 80$            & $-$     & $-$        & $-$   \\
            Encoder 10           & $-$  & $120, 80$           & $-$     & $-$        & $-$   \\
            Decoder 1            & $-$  & $-$                 & $64$    & $64$       & $-$   \\
            Decoder 2            & $-$  & $-$                 & $40$    & $40$       & $-$   \\
            Enhancement 1        & $-$  & $40, 40$            & $-$     & $-$        & $-$   \\
            Enhancement 2        & $-$  & $40, 40$            & $-$     & $-$        & $-$   \\
            Decoder 3            & $-$  & $-$                 & $16$    & $-$        & $-$   \\
            Final Block          & $-$  & $-$                 & $-$     & $-$        & $2$   \\
            \bottomrule
        \end{tabular}
    \end{center}
    \caption{
        The number of channels in different components of the proposed network.
    }
    \label{tab:model-channels}
\end{table*}

Table~\ref{tab:model-channels} illustrates the number of channels used in each component of MMNet.
The initial block outputs a 32 channel feature map, as described in the first row.
The numbers in the encoder/enhancement columns represent the number of channels returned by the multi-branch $1 \times 1$ convolutions and the final output of the encoder/enhancement block after the concatenation.
For example, encoder \#6 will receive a $40$ channel input which the $1 \times 1$ convolutions in multiple branches each expand to $64$ channels.
After the multi-branch, the outputs are concatenated and convoled by a $1 \times 1$ convolution which compresses the number of channels back to $40$.
Whenever there is a skip connection, the output of a decoder block is concatenated with the output of a refinement block.
Their respective number of channels are delineated in the decoder rows.
The final block returns a two-channel output, each for foreground and background.

\end{appendices}

\end{document}